\documentclass{sigkddExp}
\usepackage{multicol,lipsum}

\begin{document}
%

\title{
Deep Expectation-Maximization for Semi-Supervised \\ Lung Cancer Screening\\
}
%

%



\author{Sumeet Menon$^1$, David Chapman$^1$, Phuong Nguyen$^1$, Yelena Yesha$^1$, \\ \\
Michael Morris$^1$, Babak Saboury$^2$ \\
\\ 
\normalsize{$^1$University of Maryland, Baltimore County \quad \quad \quad \quad \quad $^2$University of Maryland, Medical Center}\\
\normalsize{1000 Hilltop Circle, Baltimore, MD, 21250 \quad \quad \quad \quad \quad \quad 22 S Greene St, Baltimore, MD 21201}\\
\normalsize{sumeet1@umbc.edu} \quad \quad \quad \quad \quad \quad \quad \quad \quad \quad \quad \quad \quad \quad \quad \quad \quad \quad \quad \quad \quad \quad \quad \quad}


\maketitle

\begin{abstract}
We present a semi-supervised algorithm for lung cancer screening in which a 3D Convolutional Neural Network (CNN) is trained using the Expectation Maximization (EM) meta-algorithm. Semi-supervised learning allows a smaller labeled data-set to be combined with an unlabeled data-set in order to provide a larger and more diverse training sample. EM allows the algorithm to simultaneously calculate a maximum likelihood estimate of the CNN training coefficients along with the labels for the unlabeled training set which are defined as a latent variable space. 
We evaluate the model performance of the Semi-Supervised EM algorithm for CNNs through cross-domain training of the Kaggle Data Science Bowl 2017 (Kaggle17) data-set with the National Lung Screening Trial (NLST) data-set.  Our results show that the Semi-Supervised EM algorithm greatly improves the classification accuracy of the cross-domain lung cancer screening, although results are lower than a fully supervised approach with the advantage of additional labeled data from the unsupervised sample.  As such, we demonstrate that Semi-Supervised EM is a valuable technique to improve the accuracy of lung cancer screening models using 3D CNNs.  
\end{abstract}

\section{Introduction}
The accuracy of Computer Aided Diagnosis (CAD) for cancer screening has improved tremendously in recent years due to advances in Deep Learning\cite{b1, b2, b3, b5, b10, b13, b14}. The most successful deep learning algorithm to date for image classification being the Convolutional Neural Network (CNN).  However, a major limitation of most deep learning algorithms including CNNs is that they are fully supervised.  As such, low data volumes of labeled imagery are often limiting factor, especially in the medical imaging domain in which accurate data labels require a great deal of clinical training to be able to construct.

Semi-supervised learning is an approach to expand the volume and diversity of the labeled training sample set by making use of an additional unlabeled training sample. This approach can increase data volumes thereby potentially improving screening accuracy. However, semi-supervised learning introduces additional complexity into the training process.


\vspace{10pt}
Expectation maximization (EM) is a classical statistical meta-algorithm for estimating a model given some variables existing in a latent variable space \cite{b18}. Although in machine learning EM is often associated with relatively simple Gaussian Mixture Models (GMM), it can also be applied to more complicated deep learning models including CNNs thereby enabling CNNs to be trained in the presence of latent variables. In our approach, we perform semi-supervised learning by employing the EM meta-algorithm to train the maximum likelihood CNN model in the presence of both observed and unobserved (latent) image labels.

For training and evaluation we combined two lung cancer screening data-sets: The Kaggle Data Science Bowl 2017 (Kaggle17) as well as the National Lung Screening Trial (NLST). Kaggle17 has a total of 1375 patients and the Computed Tomography (CT) scans include image volumes with associated binary clinical labels for 365 patients diagnosed with lung cancer within one year of the scan. The CT-scans for which the patient was diagnosed with lung cancer are labelled as 1 and the remaining are labelled as 0. NLST is another data-set we have used for training our model. The subset of the NLST made available to us for this model contained 4075 patients out of which 639 patients had been diagnosed with lung cancer. 

Combining two cancer screening data-sets in this way introduces potential challenges with cross-domain training.  Even though both data-sets represent imagery from a similar lung cancer screening task, there are small discrepancies that may impact the classification performance.  Furthermore, the NLST data-set is 4x larger than the Kaggle17.  As such, we want to evaluate the ability to train a model using supervised imagery from one data-set and to incorporate unsupervised imagery from another data-set to perform semi-supervised learning.  In this study, we demonstrate that a semi-supervised training approach using EM is able to achieve greater accuracy than a fully supervised approach using either the NLST or Kaggle17 data-set on it's own.  
We find that semi-supervised learning with EM is able to increase the available labeled data volume and thereby improve the accuracy of deep cancer screening tasks.



\section{Related Work}
Semi-supervised learning comprises a variety of methods to combine labeled and unlabeled training data to improve model performance \cite{b1, b7, b8, b9, b12}. 
Generative techniques attempt to model the probability distribution of the unlabeled imagery as a function of the model and labeled imagery \cite{b19, b20}.  The most influential generative technique is Expectation Maximization introduced by Dempster et. al 1977 \cite{b17}.  EM is originally intended to be applied to generative models although many important classifiers including CNNs are of the discrimination variety.  Nevertheless, due to their ability to infer label probabilities from source imagery, it is possible to make the assumption that the discrimination model is approximately generative \cite{b21}.  Papandreou et. al (2015) has combined EM with CNNs to infer pixel segmentation from weak image labels \cite{b22} demonstrating that EM is compatible with CNNs.  
There are also methods that use the labels predicted in the first iteration that combine with the multiple machine learning models through ensemble methods \cite{b7} called Co-Training. Active learning has also been combined with Expectation Maximization \cite{b8} which is a related method that helps human annotators guide the Semi-Supervised learning by the machine selecting data-points to fully label. In this scenario, the model constantly trains on new data and labels being fed in every iteration to classify the unlabelled data. 
Finally, extensions to EM have been incorporated into a variety of shallow classifiers including SVMs and HMMs.  These alternative techniques include Co-Training and Co-EM which introduce additional non-linearity into Semi-Supervised learning by incorporating multiple views of the unlabeled imagery \cite{b12}.


\vspace{0.7cm}

\section{Data Preparation}
\begin{figure}[htbp]
\centerline{\includegraphics[width=1\columnwidth]{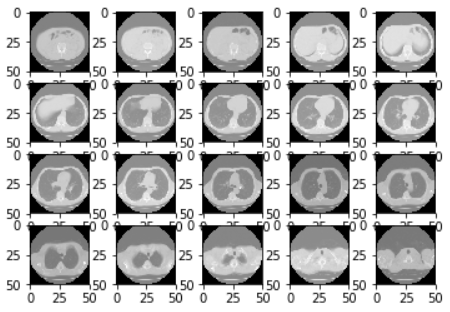}}
\caption{Axial slices of lung CT-scan subsequent to pre-processing.}
\label{slices}
\end{figure}

Both the Kaggle17 and NLST datasets contain chest CT scans with slice thickness less than 3mm. These CT scans are a 3D volume with every voxel value having a single intensity value in accordance with the Hounsfield scale's standardized units. Data pre-processing was performed prior to model training. Each axial slice of the data-set is $512\times512$ and the number of axial slices per CT scan varied between 150 to 225 in each volume. We have created chunks of 20 slices for every patient and re-sized each image to $50\times50$. The suspicious nodules included in the data-set are on the order of 1 cm\textsuperscript{3}, slices from an example CT image are shown in figure \ref{slices}

\vspace{0.7cm}

\section{CNN Architecture}

\begin{figure}[htbp]
\centerline{\includegraphics[width=1\columnwidth]{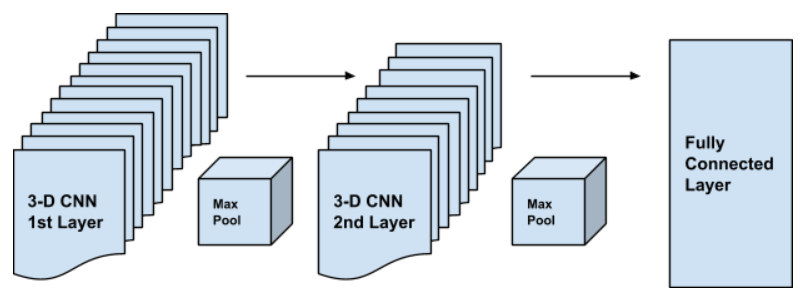}}
\caption{Architectural diagram of 2 layer 3-D CNN model.}
\label{2_Layer_CNN}
\end{figure}

\begin{table}[htbp]
\caption{2 layer 3-D CNN model}
\begin{center}
\begin{tabular}{|c|c|c|}
\hline

\textbf{Layer Number} & \textbf{\textit{Name}}& \textbf{\textit{Output Shape}} \\
\hline
\textbf{0}& Input& 50  \\
\hline
\textbf{1}& 3-D CNN& 32  \\
\hline
\textbf{2}& Max-Pooling& 32\\
\hline
\textbf{3}& 3D-CNN& 64   \\
\hline
\textbf{4}& Max-Pooling& 64   \\
\hline
\textbf{5}& Fully Connected Layer& 1024  \\
\hline

\end{tabular}
\label{tab1}
\end{center}
\end{table}

\begin{figure}[htbp]
\centerline{\includegraphics[width=1\columnwidth]{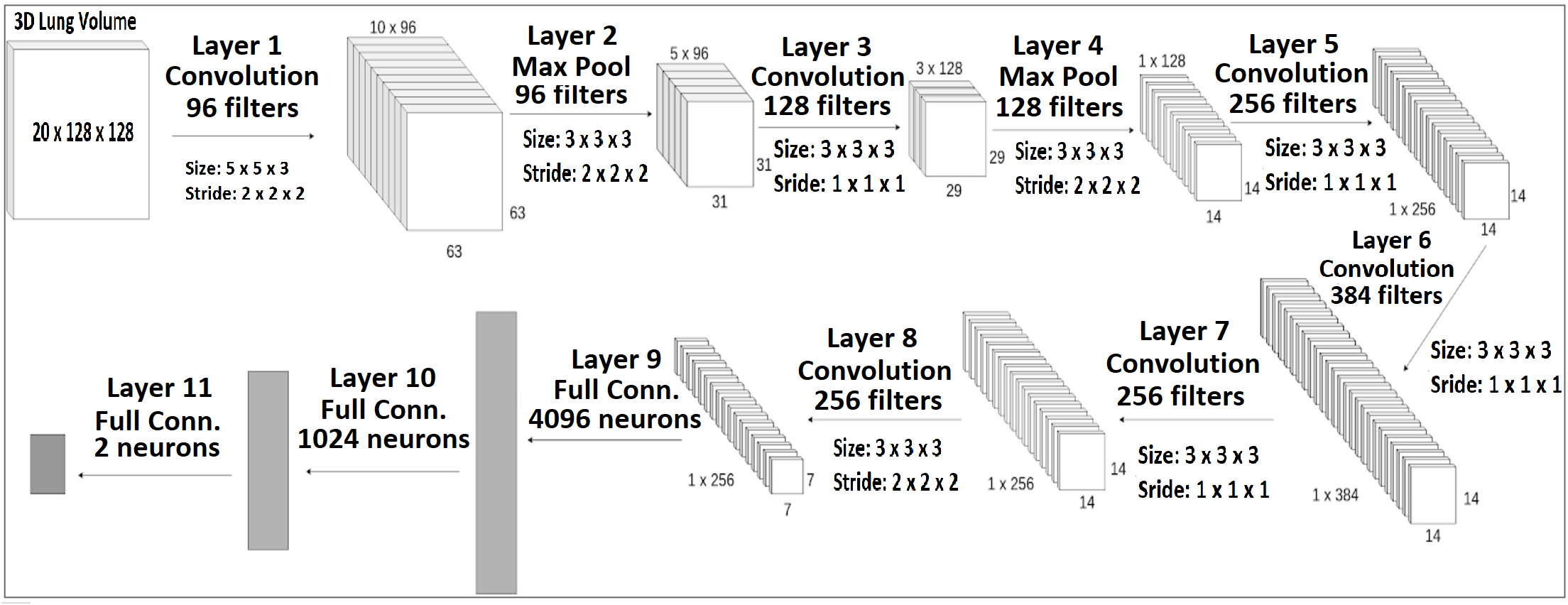}}
\caption{Architectural diagram of 3D AlexNet CNN.}
\label{AlexNet}
\end{figure}

We evaluate the Semi-Supervised EM method using two very different CNN architectures for lung cancer screening. The first architecture is a relatively simple 2-layer 3D CNN as seen in Figure \ref{2_Layer_CNN}, and the other is a deeper 3D AlexNet architecture as seen in Figure \ref{AlexNet}.

The architecture of our 3D 2-layer CNN in Figure \ref{2_Layer_CNN} consists of 2 layers of CNN with a fully connected dense layer in the end. We have passed a sliding $3\times3$ window over each 3-D image for feature extraction. In the first layer of the CNN, the sliding window generates features and passes it to the max pooling layer which reduces the size of the feature maps before passing it to the next convolution layer.

AlexNet comprises of 11 layers which are a combination of convolutional, max pooling, and fully connected layers. The detailed description is illustrated in the figure \ref{AlexNet}.

\subsection{2 layer 3D - CNN}

In this model the 3D volume data is first passed through the 3D-CNN layer with 1 channel where it detects 32 features and is passed to the next max-pooling layer. It computes the first 32 features using a window of size $3\times3\times3$. In the max-pooling layer, it computes the highest pixel-values and creates a new image. In the next 3D-CNN layer it computes 64 features using 32 channels and is then passed on to the next max-pooling layer. Finally, it is passed through the fully connected layer which computes 1024 features. The architectural diagram for this model can be seen in Figure \ref{2_Layer_CNN} and output shape in Table 1.

\subsection{AlexNet}

In this model as shown in Figure \ref{AlexNet}, the 3-D Lung volume is passed through the first layer of CNN with 96 filters, size being $5\times5\times3$ with a stride of $2\times2\times2$ and setting and zero padding. It is then passed on to the next layer of max-pooling with the same 96 filters with the size as $3\times3\times3$ and stride as $2\times2\times2$. In the next layer of CNN the filters are increased to 128 keeping the same size but reducing the stride to $1\times1\times1$. In the 4\textsuperscript{th} layer during max-pooling the strides and the size is unchanged. The next layer of CNN has 256 filters with size $3\times3\times3$ and stride $1\times1\times1$ but with no padding. The 6\textsuperscript{th} layer is a CNN with 384 filters with the same size and stride and zero padding is added. The 7th layer again is a CNN layer with 256 filters with size 3x3x3 and the stride as $2\times2\times2$. The 9\textsuperscript{th},10\textsuperscript{th} and 11\textsuperscript{th} layers then consists of 4096, 1024 and 2 neurons respectively. These are fully connected layers.





\vspace{0.7cm}
\section{Expectation Maximization}

The EM algorithm is used to find the maximum-likelihood of a model in the presence of observed and latent variables.  
It is a technique that can be used in a semi-supervised approach to infer unknown labels while training.  In our method, based on the labelled data-points, EM initially generates a classifier $\theta$. The next step consists of performing an iterative procedure where EM uses $\theta$ to classify the data and then generate a new MAP hypothesis based on the labels inferred in the previous step. 

EM is an iterative method that attempts to determine the latent variable $Z$ which in our case are a set of unknown image labels, such as to maximize the likelihood of observing the image $X$ given a CNN model.  The likelihood of a latent variable is given by the integral of the joint probability density over all possible values of the latent variable $Z$.

\begin{equation}
    L(\theta ; X) = p( X | \theta ) = \int p(X,Z | \theta) dZ
\end{equation}

EM attempts to solve the above integral by alternating between Expectation and Maximization steps. Expectation is in which we calculate the expected value of the latent variables given the t\textsuperscript{th} iteration of the model $\theta^t$.  In the context of a deep learning framework, the expected value of $E_{Z|X,\theta^t}$ can be computed by classifying label probabilities of the unlabeled imagery using the t\textsuperscript{th} iteration of the model coefficients $\theta^t$.

\begin{equation}
    Q(\theta | \theta_t) = E_{Z|X,\theta^t}[log L(\theta;X,Z)]
\end{equation}

The Maximization step is to compute the maximum likelihood model $\theta^{t+1}$ given our current expected value of the latent variables $Z$.  This can be accomplished by retraining the deep learning model using the expected value of the image labels at the t\textsuperscript{th} iteration.

\begin{equation}
    \theta^{t+1} = \text{argmax}_\theta Q(\theta | \theta^t )
\end{equation}
  
Our algorithm is similar to Co-EM \cite{b12} which is a semi-supervised algorithm which makes use of the hypothesis learned in one view to make use of probability to label the examples in the other data-set. It runs EM in every iteration and interchanges the labels that it has learned using probability. The major difference with this algorithm is that it does not set the labels in one iteration but based on the probability, it changes the labels and gets it close to the cluster it belongs to.

There are several variants of EM for semi-supervised learning that evaluate this system in slightly different ways. In order to improve convergence, we decided to incorporate the unlabeled imagery gradually over multiple iterations of EM rather than all at once.  When training on NLST and evaluating on Kaggle17, in the first iteration we predict the labels of the first 200 images in the Kaggle17 data-set. After prediction of the labels we concatenate these labels to the respective images and append it to the training data-set to pass it through the model again to predict the labels of the next 200 images in the next iteration. We repeat this process until the likelihood of the labels are maximized and the labels of all the images in the unknown data-set are predicted.



\section{Evaluation}

We evaluate the accuracy of the Semi-Supervised EM methodology for cross-domain lung cancer screening.  We compare our results to a fully supervised baseline as well as a fully supervised upper-bound.  To perform this evaluation, we combined the NLST and Kaggle17 data-sets in several ways as described in tables 2 and 3.  

In both tables we wish to evaluate the ability to train the CNN model on one Lung Cancer data-set (either Kaggle17 or NLST) and evaluate classification accuracy on the other data-set. This task is cross-domain in the weak sense that the Kaggle17 and NLST datasets are highly related but differences can affect classification accuracy. In Table 2 we train the CNN models using supervised data from the Kaggle17 data-set and evaluate the accuracy for cancer screening on the NLST data-set.  Respectively, in Table 3 we train the CNN models using supervised data from NLST and evaluate using Kaggle17.

The question we wish to answer is to what extent we are able to improve the accuracy of this cross-domain classification task by incorporating unlabeled data from the evaluation domain's training set using Semi-Supervised EM.  As such, both Tables 2 and 3 have three columns.  The first column shows a baseline fully supervised approach using only the labeled out-of-domain data-set for training.  The second column shows a fully supervised upper-bound of using both in-domain and out-of-domain data for training.  The third column "Semi-Supervised EM" shows the extent to which including unlabeled data from the evaluation domain using the proposed methodology is able to improve the classification performance.

Table 2 shows the performance of supervised baseline, supervised upper-bound, and Semi-Supervised EM to classify NLST imagery using supervised imagery from Kaggle17. We divide the training, validation and test data into 80, 10 and 10 percent respectively. The 2 layer 3-D CNN model gives us baseline accuracy of 75.95\% which improves to 77.5\% with the addition of unlabeled imagery from NLST.  Similarly the AlexNet gives us a baseline accuracy of 79.36\% which improves to 81.1\% using Semi-Supervised EM. We see that for both CNN architectures, the improvement of incorporating unlabeled imagery from NLST with Semi-Supervised EM is roughly half that of an upper-bound using fully labeled imagery from both data-sets combined.


Table 3 shows a similar cross-domain evaluation of supervised baseline, supervised upper-bound, and Semi-Supervised EM but this time to classify Kaggle17 imagery using labeled imagery from NLST.  In this process, we have trained the data-set on 80\% of the NLST data-set, validated on 10\% of the NLST data-set and tested the accuracy of the model on 10\% of the Kaggle17 data-set. The 2 layer 3-D CNN model gives us baseline accuracy of 76.75\% which improves to 78.1\% with the addition of unlabeled imagery from NLST.  The AlexNet, however shows a improvement in this case with a labeled baseline accuracy of 72.9\% which improves to 74.4\% using Semi-Supervised EM.  The NLST data-set is much larger than the Kaggle17 data-set with 4075 and 1375 CT scans respectively. The larger improvement of the 2-layer 3D CNN relative to the AlexNet can be explained because in Table 3 the unlabeled imagery is a smaller fraction of the overall data volume relative to the experiment in Table 2.

We see in both of our cross-domain lung cancer screening experiments that the Semi-Supervised EM (column 3) is able to improve the classification accuracy over a baseline supervised algorithm using only the out-of-domain imagery and labels. Also, as expected we also see that the classification accuracy is less than the upper-bound of incorporated fully supervised labels from both data-sets. Furthermore, we see that this improvement is more pronounced in the experiment of Table 2, in which we use labeled imagery from the smaller Kaggle17 data-set and incorporate unlabeled imagery from the larger NLST data-set.

\begin{table}[htbp]
\caption{Train:Kaggle17 Test:NLST}
\label{Tab_Kag_NLST}
\begin{center}
\begin{tabular}{|p{1.7cm}|p{1.6cm}|p{1.6cm}|p{1.9cm}|}
\hline

\textit{\textit{Train - Test}} & \textit{\textit{(Supervised) Kaggle17 only }}& \textit{\textit{(Supervised) Kaggle17 + NLST}}& \textbf{\textbf{Semi-Supervised EM}} \\
\hline
\textit{(2 layer 3D-CNN) NLST}& 75.95\%& 80.37\% &\textbf{77.5\%}  \\
\hline
\textit{\textit{(AlexNet) NLST}}& 79.36\%& 83\% &\textbf{81.1\%}  \\
\hline

\end{tabular}
\label{tab1}
\end{center}
\end{table}

\begin{table}[htbp]
\label{Tab_NLST_Kag}
\caption{Train:NLST Test:Kaggle17}
\begin{center}
\begin{tabular}{|p{1.9cm}|p{1.6cm}|p{1.6cm}|p{1.9cm}|}
\hline

\textit{\textit{Train - Test}} & \textit{\textit{(Supervised) NLST only }}& \textit{\textit{(Supervised) NLST + Kaggle17}}& \textbf{\textbf{Semi-Supervised EM}} \\
\hline
\textit{\textit{(2 layer 3D-CNN) Kaggle17}}& 76.75\%& 81\% &\textbf{78.1\%}  \\
\hline
\textit{\textit{(AlexNet) Kaggle17}}& 72.9\%& 77.7\% &\textbf{74.4\%}  \\
\hline

\end{tabular}
\label{tab1}
\end{center}
\end{table}





\vspace{10pt}
\section{Conclusion}

We have demonstrated that Semi-Supervised EM, when applied to computer-aided lung cancer screening with CNN models, is able to increase accuracy of cross-domain classification by incorporating unlabeled imagery.  Semi-Supervised EM is a technique to infer the maximum likelihood CNN coefficients in the presence of labeled and unlabeled imagery.  This technique can therefore improve and increase the availability of training data which is often a limiting factor for cancer screening applications.  Our findings show that using labeled imagery with the smaller Kaggle17 data-set and incorporating unlabeled imagery from the larger NLST data-set provides roughly half of the accuracy benefit as incorporating fully labeled imagery from NLST. We believe that semi-supervised learning could have a major impact on the performance of Deep CAD algorithms which are an area of active research. These results obtained help us to indicate that Semi-Supervised EM is an appropriate methodology for these purposes, and is compatible with the genre of CNN architectures in active research and development for oncology detection and diagnosis applications.

\section{Acknowledgements}

We would like to thank National Science Foundation's Center for Accelerated Real Time Analytics for supporting this project. We would also like to thank Networking health for discussing the clinical aspects of the project with us. 

\end{document}